%% file: main.tex
\documentclass[10pt,twocolumn,letterpaper]{article}

\usepackage[pagenumbers]{cvpr} 


\usepackage[normalem]{ulem}
\usepackage{multirow}
\usepackage{pifont}
\newcommand{\cmark}{\ding{51}}%
\newcommand{\xmark}{\ding{55}}%
\usepackage{makecell}
\usepackage{booktabs}
\usepackage{subcaption}
\usepackage{color,soul}

\definecolor{cvprblue}{rgb}{0.21,0.49,0.74}
\usepackage[pagebackref,breaklinks,colorlinks,allcolors=cvprblue]{hyperref}

\def\paperID{} 
\def\confName{CVPR}
\def\confYear{2026}

\title{SENSE: \uline{S}tereo Op\uline{EN} Vocabulary \uline{SE}mantic Segmentation}

\author{Thomas Campagnolo\textsuperscript{1,2} \qquad Ezio Malis\textsuperscript{1} \qquad Philippe Martinet\textsuperscript{1} \qquad Gaetan Bahl\textsuperscript{2}\\
\textsuperscript{1}Centre Inria d'Universite Cote d'Azur, France \qquad \textsuperscript{2}NXP Semiconductors, France\\
{\tt\small \{thomas.campagnolo, ezio.malis, philippe.martinet\}@inria.fr}\\
{\tt\small \{thomas.campagnolo, gaetan.bahl\}@nxp.com}
}

\begin{document}
\maketitle
\input{0_abstract}    
\input{1_introduction}
\input{2_related_work}
\input{3_method}

\input{4_experiments}
\input{5_results}

\input{6_computation_time}

\input{7_ablation}
\input{8_limitations}
\input{9_conclusion}
\input{X_suppl}

\clearpage
{
    \small
    \bibliographystyle{ieeenat_fullname}
    \bibliography{main}
}

\end{document}

%% file: 0_abstract.tex
\begin{abstract}
Open-vocabulary semantic segmentation enables models to segment objects or image regions beyond fixed class sets, offering flexibility in 
dynamic environments. However, existing methods often rely on single-view images and struggle with spatial precision, especially under 
occlusions and near object boundaries.
We propose SENSE, the first work on Stereo OpEN Vocabulary SEmantic Segmentation, which leverages stereo vision and vision-language 
models to enhance open-vocabulary semantic segmentation. By incorporating stereo image pairs, we introduce 
geometric cues that improve spatial reasoning and segmentation accuracy. 
Trained on the PhraseStereo dataset, our approach achieves strong performance in phrase-grounded tasks and demonstrates generalization in zero-shot settings. 
On PhraseStereo, we show a +2.9\% improvement in Average Precision over the baseline method and +0.76\% over the best competing method. 
SENSE also provides a relative improvement of +3.5\% mIoU on Cityscapes and +18\% on KITTI compared to the baseline work.
By jointly reasoning over semantics and geometry, SENSE supports accurate scene understanding from natural language, 
essential for autonomous robots and Intelligent Transportation Systems.

\end{abstract}

%% file: 1_introduction.tex
\section{Introduction}
\label{sec:intro}

\begin{figure}[t]
  \centering
  \includegraphics[width=\columnwidth]{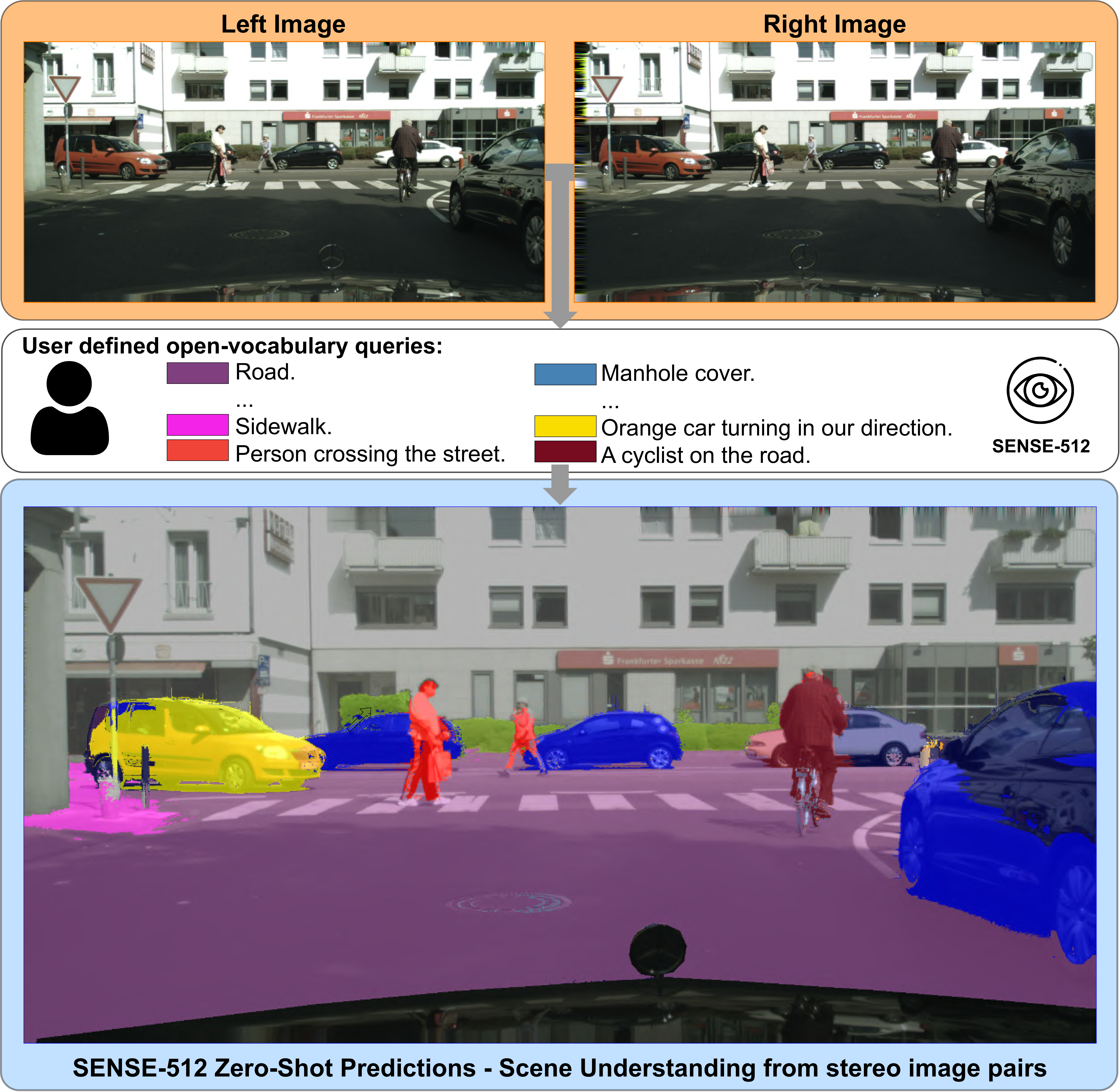}
  \caption{\textbf{SENSE-512}. Zero-shot open-vocabulary semantic segmentation from stereo image pairs (Cityscapes) using 
  natural language prompts. By combining semantics, geometry, and language, SENSE improves scene understanding 
  in Intelligent Transportation Systems (ITS).}  
  \label{fig:figure_intro}
  \vspace{-1em}
\end{figure}

Scene understanding is a fundamental task in autonomous navigation and computer vision, with semantic 
segmentation assigning class labels to every pixel. Traditional segmentation 
models~\cite{strudel2021segmenter,cheng2022masked,chen2018encoder} rely on dense annotations and operate on a 
fixed, closed set of categories. While effective in constrained settings, they struggle to generalize beyond 
predefined labels~\cite{stojnic2025lposs}.
To overcome these limitations, in the last years, open-vocabulary semantic segmentation has emerged as a promising 
alternative~\cite{akter2025image}. This paradigm enables models to segment images based on arbitrary class names or natural 
language expressions provided at inference time.
Most existing approaches rely on Vision-Language Models (VLMs) such as CLIP~\cite{radford2021learning},
which align textual and visual features to enable matching between language prompts and image content. 
However, these models are primarily designed for image-level classification and struggle with dense 
prediction~\cite{wang2025declip}, lacking the spatial granularity needed for pixel-wise segmentation.
Moreover, current open-vocabulary segmentation approaches operate on single-view 
images~\cite{luddecke2022image,zhou2022extract,zhou2023zegclip}, ignoring the geometric cues available 
in stereo vision. This limits their ability to reason about spatial relationships, which is essential 
for real-world applications such as autonomous robots and Intelligent Transportation Systems (ITS).

In this work, we introduce SENSE, the first Stereo Open-Vocabulary Semantic Segmentation method. 
SENSE leverages stereo image pairs and intermediate CLIP activations to enhance spatial reasoning and boost segmentation accuracy, 
especially near boundaries and occlusions.
Our architecture builds upon frozen CLIP features~\cite{radford2021learning} and the CLIPSeg framework~\cite{luddecke2022image}, 
adding a stereo fusion module and a lightweight decoder that jointly process intermediate representations from both views. 
Conditioning is performed in CLIP's text-image embedding space, enabling natural-language queries without retraining the backbone and 
preserving CLIP's strong generalization capabilities.

For large-scale datasets such as Cityscapes~\cite{Cordts2016Cityscapes}, we use a sliding-window strategy to address the resolution limits of CLIP encoders~\cite{radford2021learning}, 
enabling SENSE to produce fine-grained predictions while preserving global context.
In the zero-shot setting, per-prompt outputs are converted to multi-label predictions using CRF refinement~\cite{lafferty2001conditional}.
Overall, SENSE achieves competitive performance across zero-shot and referring expression segmentation, and generalizes well to unseen categories and natural language queries.

Our contributions are summarized as follows:
\begin{itemize}
  \item We introduce SENSE, a novel stereo-based architecture for natural language and open-vocabulary semantic segmentation.
  \item We propose a fusion mechanism that integrates stereo CLIP features with prompt-conditioned segmentation decoding.
  \item We train SENSE on the PhraseStereo dataset~\cite{campagnolo2025phrasestereo}, demonstrating the ability of our architecture to generalize to free-form natural language text. 
  \item We show that SENSE achieves strong generalization and competitive performance across multiple segmentation tasks, including referring expression segmentation and zero-shot multi-class semantic segmentation.
\end{itemize}

%% file: 2_related_work.tex
\section{Related Work}
\label{sec:rel_work}

\paragraph{VLMs and Semantic Segmentation.}
Self-supervised vision models have shown strong object localization capabilities~\cite{caron2021emerging,darcet2024vision,oquab2024dinov}, 
motivating efforts to enhance Vision-Language Models (VLMs) for open-vocabulary semantic segmentation~\cite{wysoczanska2024clip,lan2024proxyclip}.
VLMs enable segmentation from free-form natural language 
prompts, moving beyond fixed label sets. In ITS, this allows 
autonomous vehicles to understand from text queries, e.g., \textit{"a dangerous object in the middle of the road"} or 
\textit{"the crosswalk next to the corner"}, enhancing adaptability in complex environments~\cite{zou2023segment}.
Recent advances in VLMs and semantic segmentation have made this modality fusion increasingly effective, 
offering semantic richness and contextual flexibility~\cite{zhang2024vision}. 
VLM-based segmentation pipelines typically consist of a visual encoder (e.g., ResNet~\cite{he2016deep}, ViTs~\cite{zou2023segment}), a 
language encoder (e.g., BERT~\cite{devlin2019bert}), and a multimodal fusion module that aligns features to produce fine-grained masks~\cite{liu2024grounding}. 
Modern open-vocabulary segmentation frameworks also incorporate task-specific heads to generate binary or multi-class masks 
aligned with the provided language description~\cite{akter2025image}.

\vspace{-1em}
\paragraph{CLIP-based semantic segmentation.} Recent studies have shown that segmentation masks can be derived directly from attention maps or 
internal representations of models such as CLIP~\cite{radford2021learning}, or DINO~\cite{caron2021emerging}, even though these models were not originally 
designed for segmentation. CLIP, in particular, was trained on 400 million image-text pairs collected from the internet,
learning to align visual and textual modalities in a shared embedding space~\cite{radford2021learning}. 
This contrastive learning paradigm enables a wide range of vision tasks without task-specific fine-tuning.
CLIP's architecture and pretrained embeddings laid the foundation for various 
open-vocabulary semantic segmentation models. For instance, CLIPSeg~\cite{luddecke2022image} showed that CLIP features can be adapted for dense prediction 
with a lightweight decoder. The model is capable of segmenting images based on any text query or example image.
MaskCLIP~\cite{zhou2022extract} demonstrated how segmentation knowledge is extracted from 
CLIP with minimal architectural changes, modifying the attention pooling mechanism. 
MaskCLIP+~\cite{zhou2022extract} was further used for transductive zero-shot segmentation by generating pseudo-labels for unseen classes. 
SCLIP~\cite{wang2024sclip} enhanced CLIP's segmentation potential by introducing a 
Correlative Self-Attention (CSA) mechanism, enabling dense prediction with minimal changes to the pretrained model. 
ZegCLIP~\cite{zhou2023zegclip} employed parameter-efficient prompt tuning to adapt CLIP for zero-shot segmentation.
Subsequent works such as OpenSeg~\cite{ghiasi2022scaling} combined CLIP's open-vocabulary capabilities with traditional 
segmentation backbones, while SAM~\cite{kirillov2023segment} introduced promptable segmentation at scale using geometric prompts. 
Existing vision-language segmentation methods rely solely on monocular inputs. Our proposed method, SENSE, 
extends CLIPSeg with a transformer decoder and a stereo-fusion module that integrates intermediate features and disparity, 
enabling geometry-aware, text-driven open-vocabulary segmentation.
To the best of our knowledge, no prior work has addressed this problem.

\begin{figure*}[!ht]
  \centering
  \includegraphics[width=\textwidth]{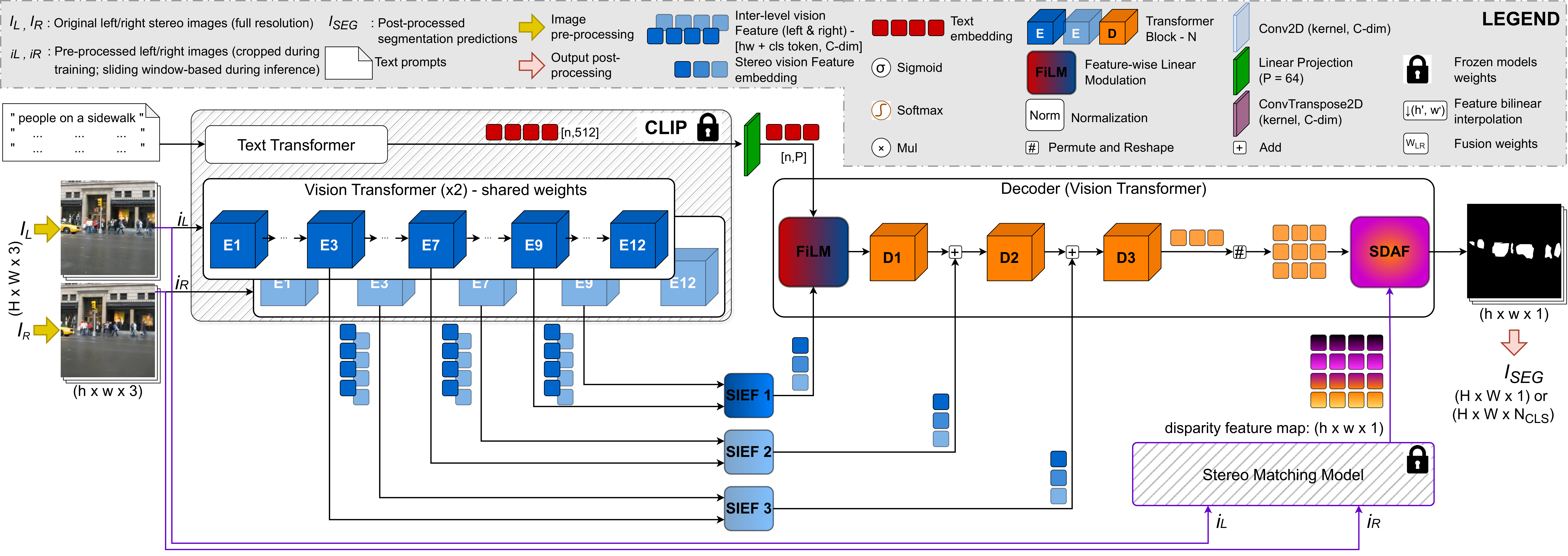}
  \caption{\textbf{SENSE architecture}. A dual-branch, weight-shared CLIP encoder~\cite{radford2021learning} processes stereo images and outputs three intermediate features per branch. These are fused via the SIEF module (\cref{sec:SIE_module}) after projection to $P=64$. The decoder comprises three transformer blocks conditioned on the textual prompt through FiLM~\cite{dumoulin2018feature}, and incorporates the SDAF module (\cref{sec:SDAF_module}) for disparity‑aware refinement. Disparity maps are provided by a frozen stereo estimator such as Selective-IGEV~\cite{wang2024selective}. The decoder, fusion modules, and projections are trained on PhraseStereo~\cite{campagnolo2025phrasestereo}.}
  \label{fig:SENSE}
  \vspace{-1em}
\end{figure*}

\vspace{-1em}
\paragraph{Referring Expression and Zero-Shot Segmentation.}
Two related tasks bridging natural language and dense prediction are Referring Expression Segmentation 
and Zero-Shot Segmentation.
Referring Expression Segmentation focuses on segmenting regions described by a natural language phrase. 
Early approaches combined recurrent language models with CNNs~\cite{kirillov2023segment,liu2017recurrent}, while 
later work introduced attention mechanisms~\cite{ye2019cross} and modular reasoning~\cite{wu2020phrasecut}.
In our work, we use the PhraseStereo dataset~\cite{campagnolo2025phrasestereo}, which extends referring expression segmentation 
to stereo images with rich object, attribute, and spatial queries.

Zero-Shot Segmentation aims to segment categories unseen during training, often suffering from bias 
toward seen classes. Early methods synthesized pixel-level features for unseen categories, 
while later approaches incorporated semantic class information~\cite{xian2019semantic} or explicitly modeled 
unseen object detection~\cite{zhang2021prototypical}.

%% file: 3_method.tex
\section{SENSE Method}
\label{sec:method}

In SENSE, we introduce a U-Net-inspired architecture with skip connections from two branches of CLIP vision encoders~\cite{radford2021learning}
to a transformer-based decoder. 
The stereo input pair is processed using the same CLIP ViT-B/16 image encoder with shared weights, 
applied independently to the left and right images. A CLIP text transformer provides the 
corresponding textual embedding.
The decoder, together with the Stereo Intermediate-level Embedding Fusion (SIEF) and Semantic Disparity Attention Fusion (SDAF) modules, is trained on the PhraseStereo
dataset~\cite{campagnolo2025phrasestereo} for open-vocabulary, natural language-driven segmentation, while all CLIP components 
remain frozen. An external stereo matching model provides disparity maps, 
which are fused with semantic features as a final-stage in the decoder.
The general architecture of SENSE is shown in \cref{fig:SENSE}.
The objective is to retain CLIP's open-vocabulary flexibility while enhancing segmentation 
accuracy, without introducing strong task-specific biases.

Embedding stereo cues directly into the shared representation lets geometry guide boundaries and resolve occlusions before mask 
prediction, producing sharper and more consistent segmentations. In contrast, monocular and post-hoc multi-view methods treat 
geometry after prediction, making them brittle to occlusions and depth discontinuities. 
This motivates integrating stereo features inside the decoder via SIEF rather than relying on post-hoc corrections.

\subsection{Encoder Architecture}

Our encoder relies on the CLIP ViT-B/16 model with a patch size of $16 \times 16$. During training, 
image preprocessing consists of normalization and cropping (from $\mathbb{R}^{ W \times H \times 3}$ to $\mathbb{R}^{ \text{w} \times \text{h} \times 3}$), 
while inference employs a sliding-window approach for processing large-scale datasets such as Cityscapes~\cite{Cordts2016Cityscapes} and KITTI 2015~\cite{Alhaija2018IJCV} (see~\cref{sec:pre_post_processing} in Supplementary Material).
The original CLIP model is constrained to a fixed input size of $224\times 224$ pixels due to its 
learned positional embeddings. Following the analysis in~\cite{luddecke2022image}, different input sizes were enabled for CLIP. 
We adopt two resolutions ($\text{w} \times \text{h}$): $352\times 352$ and $512\times 512$. 
Given a stereo image pair $\mathbb{R}^{w \times h \times 3}$, the left and right images are passed through two 
branches of the shared CLIP vision encoder (see \cref{fig:SENSE}). From each branch, we extract activations from 
transformer blocks $E = \{3, 7, 9\}$, including the CLS token.
The extracted features are fused using the Stereo Intermediate-level Embedding Fusion (SIEF) module (\cref{sec:SIE_module}), projected to the embedding size $P$, 
and integrated into the decoder through U-Net-inspired skip connections.
Text conditioning is provided by the CLIP text transformer~\cite{radford2021learning}, which generates an embedding of the query prompt. 
CLIP, which comprises the shared vision encoder and the text transformer, remains frozen during training and serves 
as a multi-modal feature backbone.

\subsection{Stereo Intermediate-level Embedding Fusion}
\label{sec:SIE_module}

\begin{figure}
  \centering
  \includegraphics[width=\columnwidth]{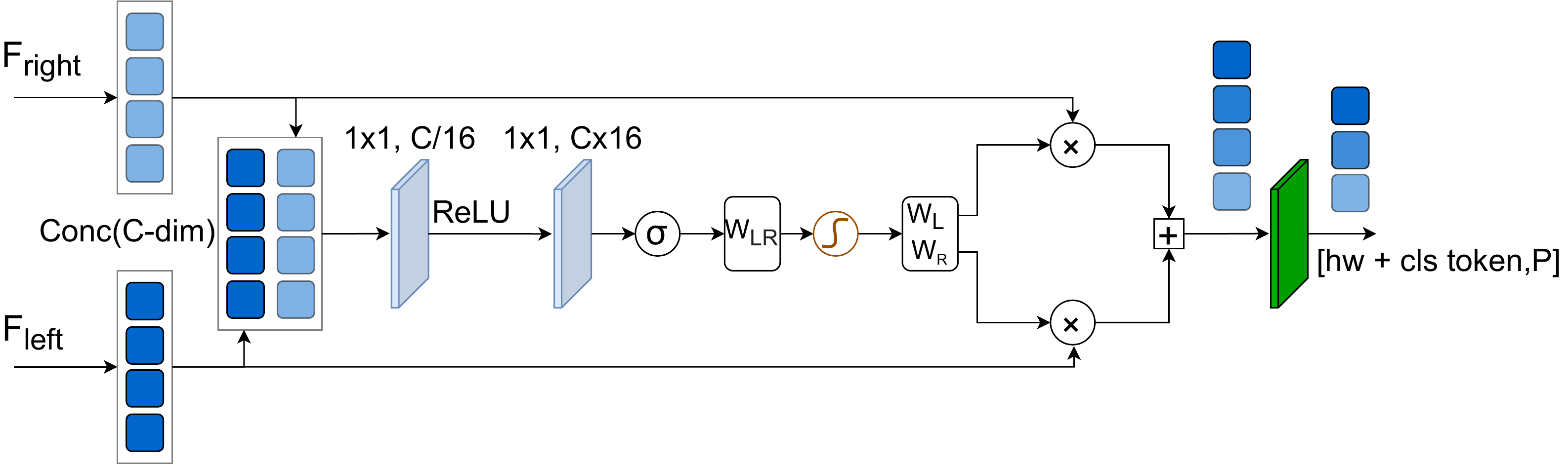}
  \caption{\textbf{Stereo Intermediate-level Embedding Fusion (SIEF) Module}. SIEF learns adaptive weights to combine
  left and right intermediate stereo activation features from CLIP vision transformer and projects the fused
  embedding to a 64-dimensional representation. For the legend symbols, refer to \cref{fig:SENSE}.}
  \label{fig:SIE_module}
  \vspace{-1.5em}
\end{figure}

To effectively combine information from the left and right image pairs, we introduce a Stereo Intermediate-level Embeddings Fusion (SIEF) module, illustrated in \cref{fig:SIE_module}. 

This module adaptively learns view-dependent fusion weights to integrate 
intermediate features from both the left $I_L$ and right $I_R$ images.
First, the left $F_{left - E}$ and right $F_{right - E}$ intermediate activations are concatenated along the channel dimension 
to form a joint stereo representation:
\begin{equation}
    F_{concat} = \text{Concat}(F_{left - E}, F_{right - E})
    \label{eq:f_concat}
\end{equation}
where $E$ denotes the activation maps extracted from the transformer blocks $\{3,7,9\}$.
This concatenated tensor is processed by a network composed of two $1 \times 1$ convolutional layers 
with an intermediate bottleneck and ReLU activation. The first $1 \times 1$ layer reduces the channel dimension 
by a factor \textit{sf} of 16 ($C/sf$), followed by a ReLU nonlinearity and a second $1 \times 1$ layer that expands the 
representation to $C \times sf$, with $sf = 16$. A sigmoid activation generates a fusion coefficient $W_{LR}$.
This process can be expressed as follows:
\begin{equation}
    W_{LR} = \sigma(\text{Conv}_{1\times 1}(\text{ReLU}(\text{Conv}_{1\times 1}(F_{concat}))))
    \label{eq:WL_WR}
\end{equation}

After computing $W_{LR}$, the tensor is split into two components $W_{L}$ and $W_{R}$ using channel-wise partitioning. 
These are then stacked and normalized with a softmax function over the channel dimension, to ensure complementary weighting between the stereo image pairs:
\begin{equation}
    [W_{L},W_{R}] = \text{Softmax}(W_{LR})
    \label{eq:WLandWR}
\end{equation}
The left feature is then fused with the right features, performing the following:
\begin{equation}
\begin{aligned}
    F_{LR} = W_L \odot F_{left} + W_R \odot F_{right}, \\
    \text{with } W_L + W_R = 1
\end{aligned}
\label{eq:Ffused}
\end{equation}
where $F_{LR}$ are the fused features from the left and right images, $\odot$ is the Hadamard product, 
and $W_L$ and $W_R$ are fusion weights for the left and right branches, respectively.
The SENSE architecture includes three Stereo Intermediate-level Embedding (SIEF 1-3) fusion modules
for progressive stereo feature integration. In addition, the mechanism allows the SIEF module to 
dynamically emphasize the informative view at each spatial location,
improving robustness to occlusions and view-specific artifacts.
Finally, the result of this fusion is projected through a linear layer (represented in green in \cref{fig:SIE_module}) 
to reduce its channel dimension to a fixed embedding size $P=64$. 
The resulting features form the input to the subsequent decoder.

\subsection{Decoder Architecture}
\label{sec:decoder}

The decoder is implemented as a transformer-based module, with the number of blocks 
matching the stereo-fused activations produced by the SIEF modules (\cref{sec:SIE_module}). In our case, this corresponds to 
three vision transformer blocks (D1-D3 in \cref{fig:SENSE}). 
Each block processes stereo-conditioned features and progressively refines the segmentation output.
The decoder predicts binary segmentation by applying a linear projection to its token representations. 
After the last transformer block D3, the fused stereo features have a token-based shape of
[w'h' + CLS token, P], where w'h' corresponds to the number of tokens for a given input resolution. 
The CLS token is removed, and the remaining tokens are reshaped into a spatial feature map of size
[w, h, P]. For example, an image input resolution of $352 \times 352$ 
with a ViT-B patch size of $16 \times 16$, w' $\times$ h' corresponds to $22 \times 22$ spatial tokens, with the CLS token already removed. 
To inform the decoder about the segmentation target, 
we apply Feature-wise Linear Modulation (FiLM)~\cite{dumoulin2018feature}, where the decoder's input activations are 
modulated by a conditional vector derived from the CLIP text transformer. 
This vector is projected to $P=64$ channels to match the fused stereo feature dimension. 
Additionally, a Semantic Disparity Attention Fusion (SDAF) module is integrated as the final stage 
of the decoder to refine segmentation using disparity maps from an external stereo matching model 
(e.g., Selective-IGEV~\cite{wang2024selective}, or any other stereo depth estimation method~\cite{liu2024one, chen2024mocha, jiang2025defom}). This module is plug-and-play and remains frozen during training. 
In the current design, the refined feature maps are upsampled to the original image resolution within the 
SDAF module. If SDAF is removed, this upsampling can be implemented as a separate final step with minor modifications 
(see~\cref{sec:sdaf_design_ablation} in the Supplementary Material for details).

\subsection{Semantic Disparity Attention Fusion (SDAF)}
\label{sec:SDAF_module}
While the decoder progressively refines stereo-conditioned 
semantic representations, fine spatial details, particularly around objects and depth can 
still be lost. To address this limitation, we introduce a Semantic Disparity Attention Fusion (SDAF) 
fusion module that leverages disparity cues as spatial attention priors to guide the 
final segmentation refinement. 
As shown in \cref{fig:SENSE}, SDAF is positioned at the end of the decoder, and directly produces 
the final segmentation prediction at the input resolution h $\times$ w.

\begin{figure}[!t]
  \centering
  \includegraphics[width=\columnwidth]{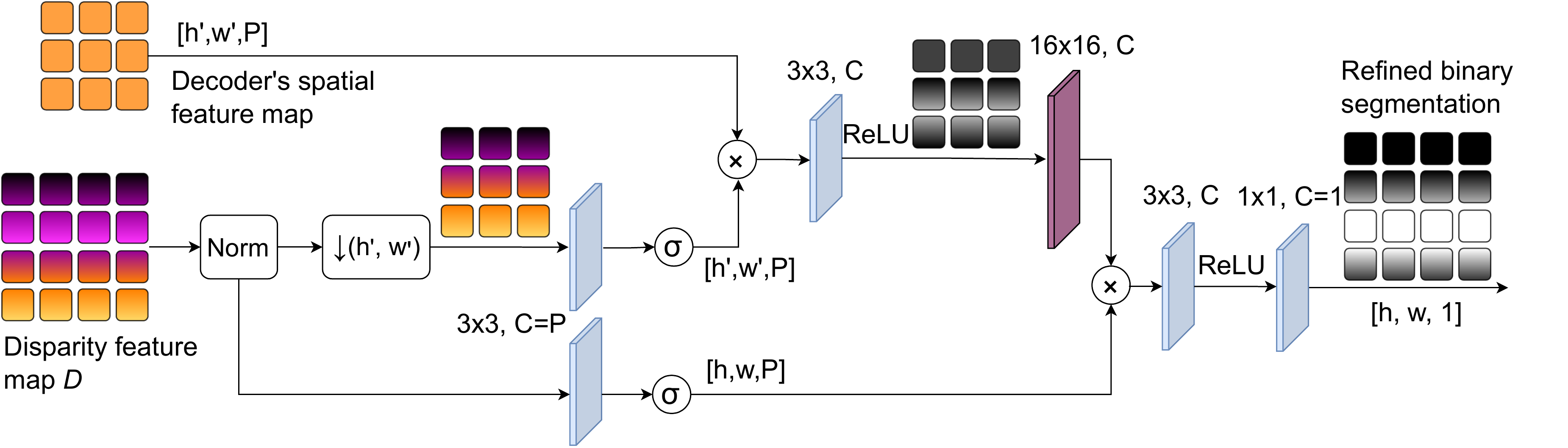}
  \caption{\textbf{Architecture of the Semantic Disparity Attention Fusion (SDAF) Module}. SDAF combines disparity-normalized
  geometric cues with semantic decoder features to refine the final segmentation output. For the legend symbols, refer to \cref{fig:SENSE}.}
  \label{fig:SDAF_module}
  \vspace{-1.5em}
\end{figure}

Represented in \cref{fig:SDAF_module}, the inputs to SDAF Module are:
\begin{enumerate}
    \item The decoder's spatial feature map $\mathbb{R}^{\text{ h'}\times\text{w'}\times P}$, obtained after the reshape of the last transformer block, and
    \item A disparity map $D \in \mathbb{R}^{352 \times 352}$ computed by an external stereo matching method, such as Selective-IGEV~\cite{wang2024selective}.
\end{enumerate}

Since the maximum disparity parameter in Selective-IGEV is set to 192.0~\cite{wang2024selective}, we standardize the 
disparity values by normalizing as $D_{norm}=D/192.0$. The normalized disparity is then 
downsampled via bicubic interpolation to match the spatial resolution (w', h') of the 
decoder feature map. This normalization and resizing process ensures that geometric cues 
are aligned and numerically consistent with the decoder's feature scale.
Next, $D_{\text{norm}}$ is processed through a sequence of convolutional layers and nonlinearities to 
generate attention-like modulation weights. These weights are applied via Hadamard product to the decoder's
spatial feature map. The modulated features are then passed through a $16 \times 16$ transposed 2D convolution, 
shown in purple in \cref{fig:SDAF_module}, which upsamples the features to the original input resolution 
$(\text{h} \times \text{w})$. Finally, a $1 \times 1$ convolution reduces the channel dimension to one, producing the 
refined binary segmentation mask.
By explicitly introducing disparity attention-like feature maps, SDAF improves spatial coherence 
near depth discontinuities and object boundaries, areas that are often challenging for purely semantic decoders. 
The stereo matching network is used as a plug-and-play, frozen component that requires 
no additional training, allowing different stereo matching architectures to be seamlessly integrated.

%% file: 4_experiments.tex
\section{Experimental setup}
\label{sec:experiments}

\paragraph{Datasets and Evaluation Metrics.}
\label{subsec:dataset_metrics}
We evaluate SENSE on PhraseStereo~\cite{campagnolo2025phrasestereo} for referring expression segmentation, 
and on Cityscapes~\cite{Cordts2016Cityscapes} and KITTI 2015~\cite{Alhaija2018IJCV}, following common open-vocabulary 
zero-shot segmentation benchmarks~\cite{akter2025image}. This setup extends evaluation to ITS-specific datasets, 
despite the limited availability of stereo datasets with semantic masks, especially for open-vocabulary segmentation.
Unlike traditional zero-shot methods with fixed label sets (e.g.,~\cite{bucher2019zero}), SENSE operates with a fully 
open vocabulary and predicts binary masks per query, adapted to a multi-label setting for ITS datasets as detailed in 
the supplementary material.

For evaluation, we report standard mean IoU (mIoU). For Referring Expression Segmentation on PhraseStereo, 
we follow prior work~\cite{luddecke2022image} and measure mIoU, foreground IoU ($IoU_{FG}$), and Average Precision (AP). 
For zero-shot semantic segmentation on Cityscapes and KITTI, we report mIoU across all classes.
It is important to note that direct comparisons can be challenging due to variations in experimental setups, 
such as the specific vocabulary used for open-set evaluation or whether the model was fine-tuned on the target dataset.

\vspace{-1em}
\paragraph{Models and Textual Prompts.}
We train and evaluate two variants of SENSE, differing in stereo input resolution. SENSE-352, which uses an input resolution of $352 \times 352$, and SENSE-512, which uses $512 \times 512$. 
This naming convention reflects the input resolution and is used consistently throughout the rest of the paper. 
Evaluating multiple resolutions lets us quantify the trade-off between efficiency and accuracy, important for ITS applications. 
Both model variants are trained on PhraseStereo, using 20\% negative samples to improve robustness to irrelevant or ambiguous 
queries, a common issue in open-vocabulary segmentation.
All reported results use a ViT-B/16 backbone, and our encoder leverages CLIP~\cite{radford2021learning} as the vision-language 
model. For referring expression segmentation, we follow the experimental setup of CLIPSeg~\cite{luddecke2022image}. 
Text prompts are taken from PhraseStereo dataset, which aligns with PhraseCut annotations~\cite{wu2020phrasecut}. 
For ITS-specific datasets~\cite{akter2025image}, which contain single-class labels such as \texttt{road}, 
\texttt{car}, etc., we use the \texttt{CLASS} name as the text prompt. This setting corresponds to an evaluation of our method in zero-shot semantic segmentation, 
since our models are trained exclusively on PhraseStereo and not on these datasets.

\begin{table*}[t]
\centering
\begin{subtable}[t]{0.49\textwidth}
\centering
\begin{tabular}{lccc}
\toprule
Method & mIoU & IoU$_{\text{FG}}$ & AP \\
\midrule
CLIPSeg$^\ast$(PC+)~\cite{luddecke2022image} & 43.7 & 55.1 & 76.7 \\
CLIPSeg$^\dagger$(PC)~\cite{luddecke2022image} & 46.1 & 56.2 & \uline{78.2} \\
MDETR$^\dagger$~\cite{kamath2021mdetr} & \textbf{53.7} & - & - \\
HulaNet$^\dagger$~\cite{wu2020phrasecut} & 41.3 & 50.8 & - \\
\\[3.0ex]
\midrule
\textbf{SENSE-352} (Ours) & \uline{47.2} & \textbf{57.0} & \textbf{78.8} \\
\textbf{SENSE-512} (Ours) & 46.5 & \uline{56.3} & 77.7 \\
\bottomrule
\end{tabular}
\caption{\textbf{Referring Expression Segmentation.}}
\label{tab:ref_expression_seg}
\end{subtable}
\hfill
\begin{subtable}[t]{0.49\textwidth}
\centering
\begin{tabular}{lcc}
\toprule
Method & Cityscapes & KITTI \\
\midrule
SAM + CLIP$^\ddagger$~\cite{akter2025image}  & 38.6 & 35.2 \\
OpenWorldSAM$^\ast$~\cite{openworldsam2025} & 23.1 & 21.1 \\
CLIPSeg$^\ddagger$(PC+)~\cite{luddecke2022image} & 34.9 & 32.1 \\
OpenSeg$^\ddagger$~\cite{ghiasi2022scaling}  & 40.2 & \uline{37.8} \\
MaskCLIP$^\S$~\cite{zhou2022extract}  & 25.0 & - \\
SCLIP$^\S$~\cite{wang2024sclip}  & 32.2 & - \\
\midrule
\textbf{SENSE-352} (Ours) & \uline{40.7} & \textbf{37.9} \\
\textbf{SENSE-512} (Ours) & \textbf{41.6} & - \\
\bottomrule
\end{tabular}
\caption{\textbf{Zero-Shot Semantic Segmentation.}}
\label{tab:zero_shot}
\end{subtable}
\caption{\textbf{Tasks performance comparison.} 
(\textbf{a}) Referring Expression Segmentation on PhraseStereo and PhraseCut (PC, PC+). 
(\textbf{b}) Zero-Shot Semantic Segmentation, in terms of mIoU, on Cityscapes and KITTI 2015.
Best results are in \textbf{bold}, second best are \uline{underlined}.
$^\ast$ indicates methods with reproduced performance, $\dagger$ denotes the numbers taken from~\cite{luddecke2022image}.
$^\ddagger$ and $^\S$ denotes the numbers taken from~\cite{akter2025image} and~\cite{stojnic2025lposs}, respectively.}
\label{tab:combined_results}
\vspace{-1em}
\end{table*}

\vspace{-1.2em}
\paragraph{Comparative Methods.}
We compare SENSE with methods in Referring Expression Segmentation. Since PhraseStereo~\cite{campagnolo2025phrasestereo} is derived from 
PhraseCut~\cite{wu2020phrasecut}, this allows us to evaluate SENSE in a stereo setting while maintaining comparability with 
existing approaches such as CLIPSeg~\cite{luddecke2022image}, MDETR~\cite{kamath2021mdetr}, and HulaNet~\cite{wu2020phrasecut}.
For Zero-Shot Semantic Segmentation, we benchmark on Cityscapes and KITTI 2015 with VLM leading 
methods, including SAM + CLIP~\cite{akter2025image}, CLIPSeg~\cite{luddecke2022image}, OpenSeg~\cite{ghiasi2022scaling}, OpenWorldSAM~\cite{openworldsam2025}, MaskCLIP~\cite{zhou2022extract}, and SCLIP~\cite{wang2024sclip}.
Our baseline selection reflects the goal of improving CLIPSeg with stereo cues, which is thus the \textit{primary} point of comparison. OpenSeg is included as the stronger available zero-shot baseline.

\vspace{-1.2em}
\paragraph{Implementation Details.}
\label{par:implementation_details}
Our method is implemented in PyTorch~\cite{paszke2017automatic}. We train the two configurations, SENSE-352 and SENSE-512.
In the Ablation Study, \cref{sec:ablation_main} and \cref{sec:ablation_supp}, we investigate the impact of replacing the CLIP vision encoder with a ResNet-50 backbone. 
Training is performed with a batch size of 64 stereo image pairs using the AdamW optimizer, a learning rate of 0.001, 
and a cosine scheduler. For CLIP feature extraction, we adopt ViT-B/16 and select the intermediate 
activation layers $E = \{3, 7, 9\}$. The fused stereo embeddings and text embeddings are projected to a dimension $P = 64$.
For disparity computation, we adopt Selective-IGEV with SceneFlow pre-trained weights~\cite{wang2024selective}. 
Trainings are conducted on the PhraseStereo dataset. All training and experiments are performed on a workstation with AMD Ryzen 9 5950X CPU, NVIDIA GeForce RTX 3090 Ti GPU, 64 GB RAM, and 2 TB NVMe SSD.

%% file: 5_results.tex
\section{Results}
\paragraph{Referring Expression Segmentation}
We present referring expression segmentation results in \cref{tab:ref_expression_seg}, evaluated using the metrics 
described in \cref{subsec:dataset_metrics} on PhraseStereo dataset. We compare our approach with transformer-based CLIPSeg~\cite{luddecke2022image}, 
HulaNet~\cite{wu2020phrasecut}, and MDETR~\cite{kamath2021mdetr}.
As shown in the table, we report two configurations of CLIPSeg trained on PhraseCut dataset~\cite{wu2020phrasecut}: (a) CLIPSeg (PC), which uses only text 
labels; and (b) CLIPSeg (PC+), which incorporates negative visual samples, similar to our training strategy described in \cref{par:implementation_details}.

Our approach, SENSE-352, surpass the two-stage HulaNet method~\cite{wu2020phrasecut} and all other baselines in terms of 
IoU$_{\text{FG}}$ and Average Precision (AP). However, for the mIoU metric, MDETR~\cite{kamath2021mdetr} achieves the best performance. 
While SENSE shows a small decrease in mIoU, this does not contradict the improvement in spatial precision 
provided by stereo cues. mIoU is heavily influenced by large, smooth background classes (e.g., \texttt{road}, \texttt{building}, \texttt{sky}), 
where monocular models such as CLIPSeg already attain high IoU. In contrast, stereo sharpens boundaries and improves 
geometric localization, which primarily benefits thin structures, occluded regions, and fine object contours, regions 
that contribute little to mIoU but strongly affect spatially sensitive metrics such as IoU$_\text{FG}$ and AP, 
where SENSE consistently performs best.
For a fair comparison, the most relevant baseline is CLIPSeg (PC+), as it matches our configuration settings (e.g., same projection dimension $P=64$), 
and it aligns with our experimental setup. SENSE outperforms CLIPSeg (PC+) across all metrics.

\begin{figure*}
  \centering
  \includegraphics[width=\textwidth]{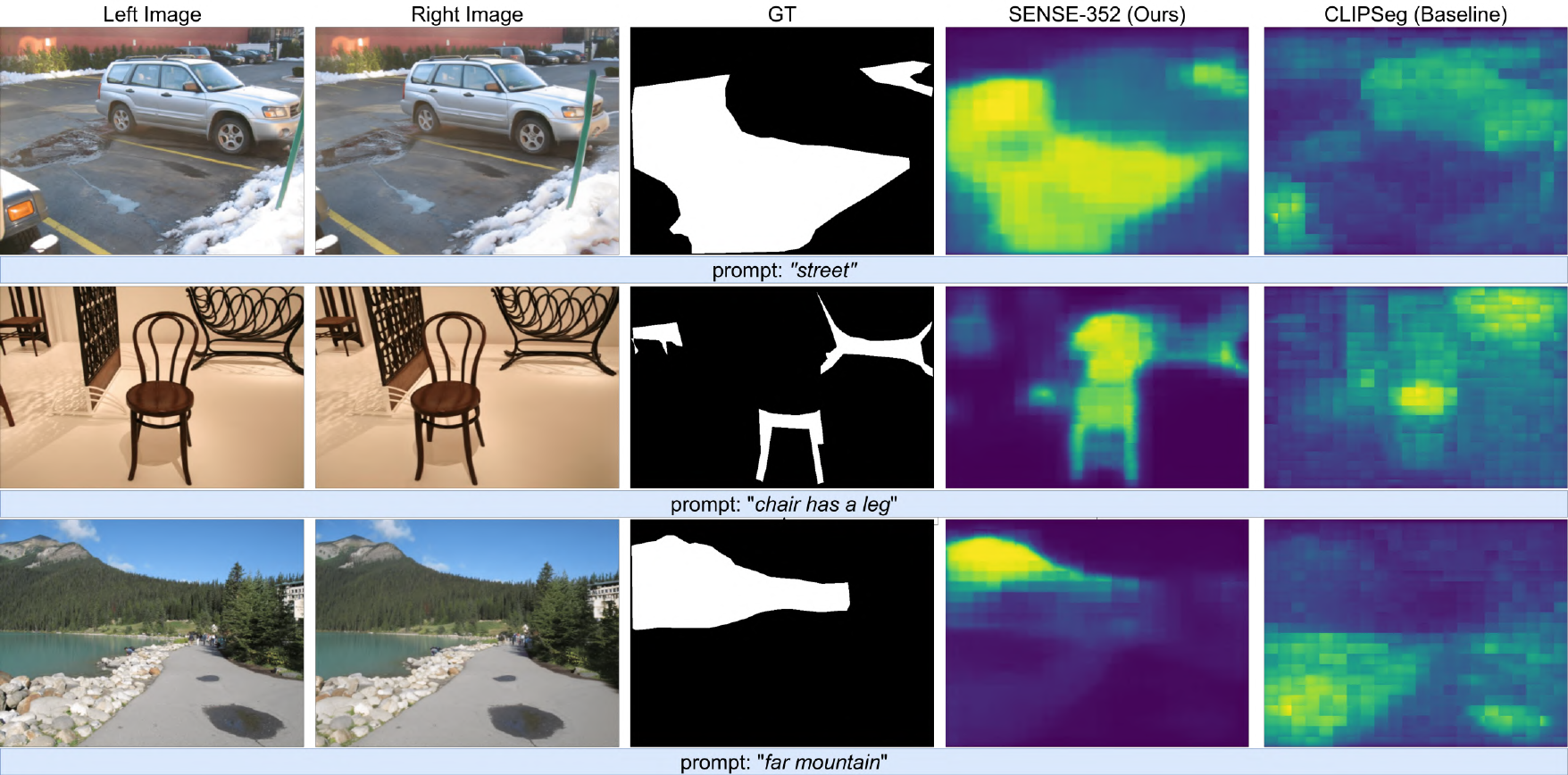}
  \caption{\textbf{Qualitative comparison of referring expression semantic segmentation.} A comparison of SENSE-352 in PhraseStereo dataset with CLIPSeg (PC+)~\cite{luddecke2022image} method. Predictions from SENSE-352 and CLIPSeg are visualized as sigmoid probability maps, highlighting confidence for the queried text prompt (blue box).}
  \vspace{-0.5em}
  \label{fig:qual_ref_expression}
\end{figure*}

Qualitative results of our method, SENSE-352, compared over the baseline CLIPSeg (PC+)~\cite{luddecke2022image}, are 
illustrated in \cref{fig:qual_ref_expression}. In the comparison, SENSE-352 produces higher probability regions that are more precisely aligned with the referred object, whereas CLIPSeg predictions appear diffuse, with probability scores spread across unrelated areas.

\paragraph{Zero-Shot Semantic Segmentation}

The results in \cref{tab:zero_shot} report zero-shot semantic segmentation performance on 
Cityscapes~\cite{Cordts2016Cityscapes} and KITTI 2015~\cite{Alhaija2018IJCV}, evaluated using the mIoU metric. 
Our approach achieves the best results on Cityscapes with SENSE-512 and on KITTI 2015 with SENSE-352, 
outperforming all other open-vocabulary baselines. 
To enable evaluation on these datasets, we apply pre- and post-processing strategies since our model was trained for binary segmentation prediction
and cannot be directly used in a multi-label setting. Specifically, we employ the sliding-window approach for 
inference and apply Conditional Random Fields (CRF) for multi-class segmentation mask refinement, as detailed in the post-processing section of the supplementary material.
Notably, SENSE-512 was not evaluated on KITTI 2015 as images of resolution $1242 \times 375$ cannot be given to a $512 \times 512$ input model without heavy padding or resizing that would distort geometry.
SENSE-352 is therefore the appropriate model variant for that dataset.

\begin{table*}[h]
\centering
\resizebox{\textwidth}{!}{
\begin{tabular}{lcccccc}
\toprule
& 
\makecell[c]{OpenSeg~\cite{ghiasi2022scaling}} &
\makecell[c]{OpenWorldSAM~\cite{openworldsam2025}} &
\makecell[c]{CLIPSeg (PC+)~\cite{luddecke2022image}} &
\makecell[c]{MDETR~\cite{kamath2021mdetr}} &
\makecell[c]{\textbf{SENSE-352} (Ours)} &
\makecell[c]{\textbf{SENSE-512} (Ours)} \\
\midrule
Time (ms)
& \textbf{123.35}
& 343.25
& 218.96
& 256.87
& \uline{194.74} (17.12)
& 284.54 (43.56) \\
\bottomrule
\end{tabular}
}
\caption{\textbf{Computational time comparison.}
We report the inference time in milliseconds (ms). Values in parentheses (...) indicate runtime \emph{without} stereo matching computation.
Best result is in \textbf{bold}, second best is \uline{underlined}.}
\label{tab:runtime_comparison}
\vspace{-1.2em}
\end{table*}

\begin{figure*}
  \centering
  \includegraphics[width=\textwidth]{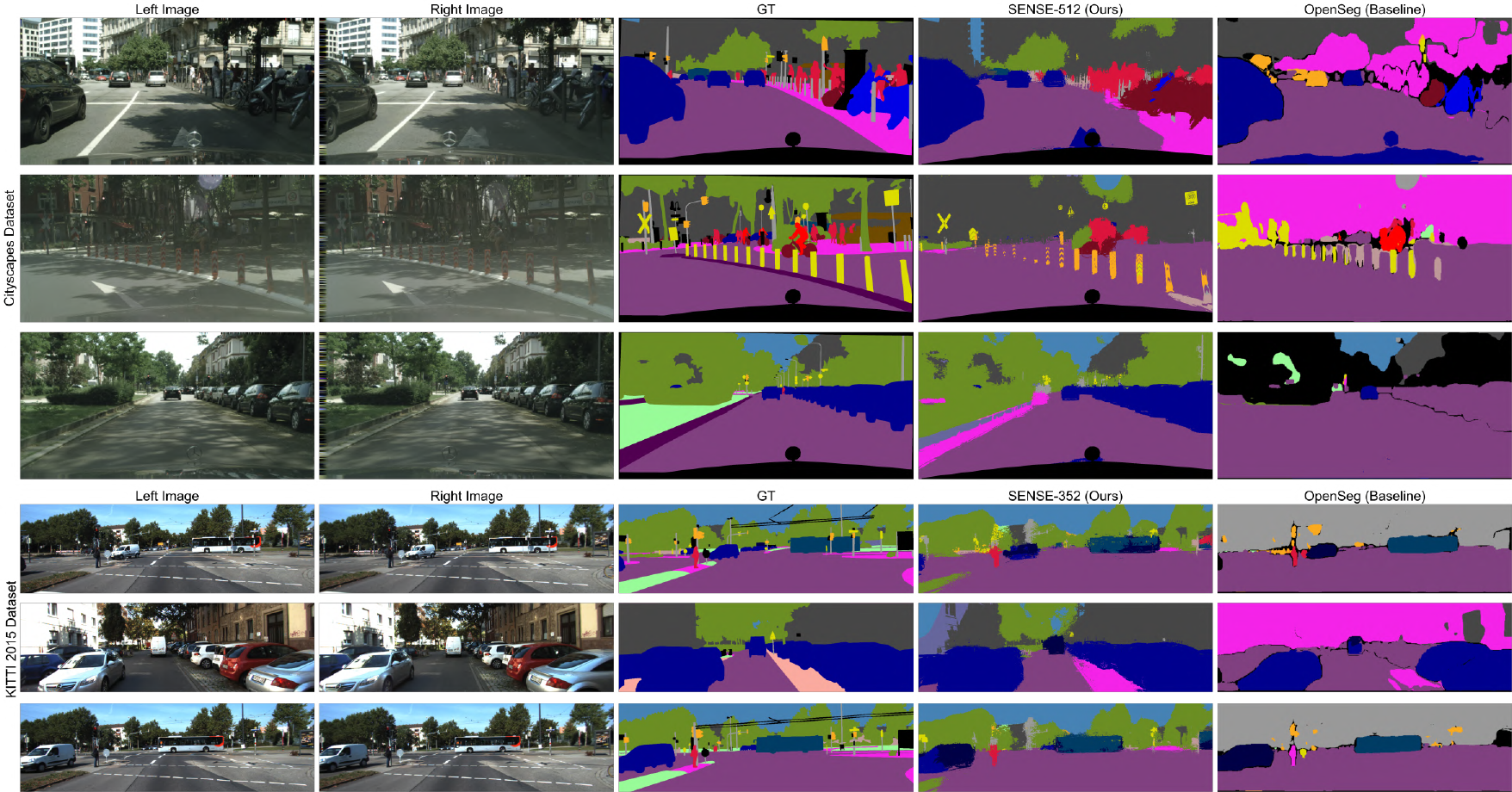}
  \caption{\textbf{Qualitative comparison of zero-shot semantic segmentation.} Comparison of SENSE-512 in Cityscapes and SENSE-352 in KITTI dataset, with the second best performing method, OpenSeg~\cite{ghiasi2022scaling}. In GT and SENSE-512, pixels shown in black are pixels that are unlabeled. In OpenSeg, black pixels correspond to \texttt{unknown} label.}
  \label{fig:qual_results}
  \vspace{-1em}
\end{figure*}

The marginal gain with respect to OpenSeg~\cite{ghiasi2022scaling} reflects the architectural differences, as our contribution
specifically enhances CLIPSeg~\cite{luddecke2022image} using the stereo formulation. Due to the unavailability of executable codebases
(e.g., OpenSeg~\cite{ghiasi2022scaling} provides only a qualitative Colab demo), we couldn't reproduce the results for all baselines,
and thus rely on the published results from the references of \cref{tab:zero_shot}.
We also compare our method against a recent state-of-the-art SAM-based approach for zero-shot open-vocabulary segmentation, OpenWorldSAM~\cite{openworldsam2025}. 
To evaluate it fairly on our datasets, we had to adapt its publicly released code. We used the referring inference pipeline with the provided RefCOCOg weights 
allowing all dataset classes. In contrast, its semantic inference script could not be used, as it relies on a COCO-specific label taxonomy that 
is incompatible with Cityscapes and KITTI benchmarks. Notably, SENSE-352 and SENSE-512 surpasses OpenWorldSAM, highlighting that our stereo 
formulation delivers substantial gains over existing SAM-based open-vocabulary approaches.
Specialized, closed-set models still outperform open-vocabulary methods in their domains, illustrating a generalization 
penalty, the performance cost of flexibility over specialization. This trade-off matters for safety-critical ITS tasks like 
lane or sidewalk detection, where specialized models remain superior. However, in scenarios requiring adaptability 
to novel concepts and richer interaction, accepting this penalty is often justified, as further discussed in \cref{sec:scene_understanding}.
We present qualitative results for SENSE-512 on Cityscapes~\cite{Cordts2016Cityscapes} and SENSE-352 on KITTI 2015~\cite{Alhaija2018IJCV} alongside leading zero-shot 
competitors such as OpenSeg (\cref{fig:qual_results}). Across diverse scenes, both SENSE variants produce outputs more aligned 
with ground truth than OpenSeg. Our sliding-window strategy combined with CRF-based refinement further improves segmentation 
quality and mitigates CLIP ViT resolution limitations. Additional results are provided in the supplementary material.

%% file: 6_computation_time.tex
\vspace{-0.4em}
\section{Computation Time}
\label{sec:computation_time}

We report inference time in \cref{tab:runtime_comparison}. Using an NVIDIA GeForce RTX 3090 Ti GPU, for a single text query, OpenSeg is the fastest baseline at 123.35 ms, 
while OpenWorldSAM is the slowest at 343.25 ms. CLIPSeg and MDETR run at 218.96 ms and 256.87 ms, 
respectively. In comparison, our SENSE models with stereo images, achieve competitive performance: SENSE-352 runs in 194.74 ms 
(and 17.12 ms without stereo matching processing), while SENSE-512 takes 284.54 ms (and 43.56 ms without disparity computation). 
This highlights that the disparity module dominates runtime and that, without it, SENSE becomes substantially faster than all compared methods.
In particular, using a real-time stereo depth estimation method~\cite{shamsafar2022mobilestereonet} would make SENSE suitable for edge deployment.

%% file: 7_ablation.tex
\section{Ablation Study}
\label{sec:ablation_main}
To identify the key components contributing to SENSE, we perform an ablation study on the PhraseStereo 
Referring Expression Segmentation task (\cref{tab:ablation}). 
All experiments use SENSE-352, which processes stereo pairs at $352\times352$ resolution.
Removing one CLIP stream together with the SIEF and SDAF modules yields the CLIPSeg~\cite{luddecke2022image} configuration, 
which serves as our monocular baseline.

The ablations show that both stereo fusion modules are essential. Removing SDAF lowers performance across all metrics, 
and using only one fusion scale is insufficient, while adding a third introduces noise, confirming the two-layer design 
as the best compromise. SIEF also provides consistent gains, as replacing its attention-based fusion with simple 
concatenation reduces accuracy. SENSE remains robust to different Selective-IGEV initializations~\cite{wang2024selective}. 
Overall, SIEF and SDAF provide complementary stereo cues, and their combination is key to SENSE's effectiveness.
For more details, please refer to~\cref{sec:ablation_supp} in the Supplementary Material.

%% file: 8_limitations.tex
\section{Discussion and limitations}

\label{sec:scene_understanding}
Traditional segmentation models perform well on closed-set labels, but real environments require adapting to novel concepts. 
Vision-Language Models are reshaping perception for autonomous systems, and SENSE 
leverages this flexibility together with stereo cues to better handle such variability. 
As shown in \cref{fig:figure_intro}, SENSE-512 can interpret rich natural-language queries (e.g., \textit{"person crossing the street"}, or \textit{"orange car turning in our direction"}) 
which are complex instructions that go beyond simple class labels. 
SENSE achieves strong results in referring expression and zero-shot segmentation, but limitations remain. 
Performance is sensitive to prompt phrasing because CLIP's text embeddings vary with wording, and negative instructions 
(e.g., \textit{“exclude cars”}) remain challenging~\cite{park2025know}. 
This highlights the need for more robust or prompt-free approaches. Additionally, prompts are currently user-defined, 
which is limited for autonomous driving. A promising direction is to automatically generate queries from scene context 
or navigation goals. Improving robustness to prompt phrasing and automating query generation is key for advancing open-vocabulary 
segmentation in dynamic environments. Another important future direction is improving computational time performance, 
essential for deployment in robotics and embedded platforms.

%% file: 9_conclusion.tex
\section{Conclusion}

We present SENSE, the first stereo-based framework for open-vocabulary semantic segmentation, a method that addresses an important gap in current 
single-view approaches. By combining stereo geometry with vision-language models, SENSE delivers improved spatial precision and robust 
zero-shot generalization across diverse benchmarks. This work underscores the potential of integrating geometry into open vocabulary semantic 
segmentation, paving the way for scene understanding in dynamic, safety-critical domains like autonomous navigation and ITS. 
We believe SENSE paves the way for future research on reliable multimodal perception and reasoning in complex real-world environments.

%% file: X_suppl.tex
\section{Appendix}
\label{sec:appendix_section}

\subsection{Ablation Study}
\label{sec:ablation_supp}
In this section, we provide an extended analysis of the ablation study reported in \cref{sec:ablation_main} of the main paper.
All experiments are conducted on the PhraseStereo Referring Expression Segmentation benchmark using the SENSE-352 configuration, 
which processes stereo pairs at a resolution of $352\times352$ pixels. Recall that removing the right-view CLIP stream together with 
both stereo fusion modules (SIEF and SDAF) reduces SENSE to the original CLIPSeg~\cite{luddecke2022image} architecture. This monocular setup, included 
in \cref{tab:ablation}, serves as the baseline against which all stereo-enabled variants are compared.

Building upon the ablations presented in the main paper, we now provide a more detailed investigation of: the design choices behind 
SDAF (\cref{sec:sdaf_design_ablation}), the impact of alternative SIEF configurations (\cref{sec:sief_design_ablation}), and the 
influence of backbone selection (CLIP ViT-B/16 vs. ResNet-50) (\cref{sec:backbone_choice_ablation}).

\vspace{-1em}
\paragraph{Effect of SDAF design.}
\label{sec:sdaf_design_ablation}
The Semantic Disparity Attention Fusion (SDAF) module, described in \cref{sec:SDAF_module} and illustrated in \cref{fig:SDAF_module}, 
is designed to fuse semantic features, calculated by the decoder, and disparity cues at multiple scales. 
We evaluate different SDAF configurations:
\begin{itemize}
  \item One-layer fusion: we simplify SDAF to a single branch, performing disparity fusion at either 22$\times$22 or 352$\times$352 resolution.
  \item Two-layers fusion (default): branches at 22$\times$22 and 352$\times$352.
  \item Three-layers fusion: we extend the default two-layer SDAF design by adding a third branch that performs disparity fusion at 88$\times$88 resolution. 
  The resulting configuration includes branches at 22$\times$22, 88$\times$88, and 352$\times$352, where the disparity map is split into three streams and 
  resized to match each scale before fusion.
\end{itemize}

For the three-scale setup, the transposed convolution layer (purple in \cref{fig:SDAF_module}) is split into two stages to progressively upsample features 
(22$\rightarrow$88$\rightarrow$352). The results show that the default two-layer SDAF configuration (indicated with $^\dagger$ in \cref{tab:ablation}) achieves the best performance. 
In contrast, one-layer fusion suffers because it lacks sufficient multi-scale context, limiting the model's ability to capture both global and 
fine-grained depth information. However, three-layer fusion introduces excessive complexity and propagates noise from disparity across scales, 
reducing accuracy. This highlights the importance of balancing multi-scale fusion without overcomplicating the design.

We also evaluate the impact of stereo matching initialization by computing disparity maps using different pretrained weights for Selective-IGEV, Middlebury vs SceneFlow,
while keeping the default two-layer SDAF configuration. Middlebury weights yield similar results, indicating robustness to stereo initialization.
Finally, removing SDAF entirely (replacing it with a simple upsampling head) significantly reduces performance, confirming SDAF's role in improving spatial coherence 
near depth discontinuities and object boundaries.

\vspace{-1em}
\paragraph{Effect of SIEF design.}
\label{sec:sief_design_ablation}
The Stereo Intermediate-level Embedding Fusion (SIEF) module, illustrated in \cref{fig:SIE_module}, integrates intermediate features from both the left image ($I_L$) and right image ($I_R$). 
Specifically, the intermediate activations $F_{left - E}$ and right $F_{right - E}$ are first concatenated along the channel dimension to form a joint stereo representation. 
This representation is then processed by an attention-based fusion mechanism to enhance cross-view consistency and exploit stereo cues.

We analyze two aspects of SIEF:
\begin{itemize}
  \item Scale factor (sf): varying $sf$ from 16 (default) to 2, which changes the channel compression ratio and the complexity of the fusion.
  \item Fusion strategy: simplifying attention-based fusion with simple concatenation of $F_{left - E}$ and right $F_{right - E}$ and 2D convolution, without any attention mechanism.
\end{itemize}

Even when SDAF is removed, enabling SIEF keeps mIoU at higher than the CLIPSeg baseline, 
showing that stereo embedding fusion alone improves the performances (\cref{tab:ablation}). 
Simplifying SIEF to a concatenation-only design results in 45.8 mIoU, meaning that 
attention-based fusion provides incremental but consistent gains. 
Varying the scale factor from 16 to 2 barely affects performance, confirming that the module is 
robust to channel compression and does not rely heavily on this parameter.

\vspace{-1em}
\paragraph{Impact of Backbone Choice.}
\label{sec:backbone_choice_ablation}
We compare ViT-B/16 and ResNet-50 as CLIP vision backbones to assess their influence on referring expression segmentation. 
As shown in \cref{tab:ablation}, ViT-B/16 consistently delivers superior results across all metrics (mIoU, IoU$_{\text{FG}}$, and AP), 
confirming that transformer-based features provide richer and more robust representations for this task. 
Nevertheless, even when using ResNet-50, SENSE-352, with the SIEF and SDAF modules, still outperforms the CLIPSeg baseline. This demonstrates the effectiveness 
of our stereo fusion design regardless of backbone choice.
The choice to keep components such as CLIP and Selective-IGEV (see \cref{sec:sdaf_design_ablation}) frozen was deliberate, 
allowing us to isolate the contribution of the stereo formulation. Jointly finetuning or replacing these components represents 
promising future work.

\begin{table*}[h]
\centering
\resizebox{\textwidth}{!}{
\begin{tabular}{lccclll} 
\toprule
Method & CLIP Image Backbone & SIEF & SDAF & \multicolumn{1}{c}{mIoU} & \multicolumn{1}{c}{$\text{IoU}_{\text{FG}}$} & \multicolumn{1}{c}{AP} \\
\midrule
CLIPSeg$^\ast$ (PC+)~\cite{luddecke2022image} & ViT-B/16 & \xmark & \xmark & 43.7 & 55.1 & 76.7 \\ 
\midrule
\multirow{8}{*}{\textbf{SENSE-352} (Ours)} & \multirow{8}{*}{ViT-B/16} 
& \cmark$^\dagger$ & \cmark$^\dagger$ &  47.2$^\dagger$ & \textbf{57.0}$^\dagger$ & \textbf{78.8}$^\dagger$ \\
& & \cmark & 3 layers & 46.7 & 56.6 & 78.0 \\
& & \cmark & 1 layer (22$\times$22) & 46.7 & 56.4 & 77.4 \\
& & \cmark & 1 layer (352$\times$352) & 45.5 & 55.9 & 77.2 \\
& & \cmark & Middlebury weights & \textbf{47.3} &  \uline{56.8} &  \uline{78.1} \\
& & \cmark & \xmark & 45.7 & 56.3 & 77.6 \\
& & $\text{sf}=2$ & \xmark & 45.7 & 56.2 & 77.8 \\
& & only Concat($F_{left-E}, F_{right-E}$) & \xmark & 45.8 & 56.1 & 77.6 \\
\midrule
CLIPSeg$^\ast$ (PC+)~\cite{luddecke2022image} & ResNet-50 & \xmark & \xmark & 40.5 & 53.1 & 75.0 \\ 
\midrule
\multirow{2}{*}{\textbf{SENSE-352} (Ours)} & \multirow{2}{*}{ResNet-50} & \cmark & \cmark & \textbf{44.1} & \textbf{54.8} & \textbf{76.7} \\
& & \cmark & \xmark & 42.6 & 54.0 & 75.9 \\
\bottomrule
\end{tabular}
}
\caption{\textbf{Ablation study conducted in Referring Expression Segmentation on PhraseStereo~\cite{campagnolo2025phrasestereo}}, using ViT-B/16 and ResNet-50 backbones.
We report mIoU, foreground IoU$_{\text{FG}}$, and Average Precision (AP).
Default setups used and explained in the paper are marked with \cmark. $^\dagger$ indicates the default configuration method.
Best and second-best results within each backbone group are shown in \textbf{bold} and \uline{underlined}, respectively. 
$^\ast$ indicates reproduced performance.}
\label{tab:ablation}
\end{table*}

\subsection{Image Pre- and Post- Processing}
\label{sec:pre_post_processing}
During inference, our method adopts a sliding-window strategy to handle high-resolution stereo images 
from large-scale datasets such as Cityscapes~\cite{Cordts2016Cityscapes} and 
KITTI 2015~\cite{Alhaija2018IJCV}. 
This approach, detailed in \cref{sec:pre_processing}, addresses the limitations 
of CLIP-based ViT encoders, which were originally trained on lower-resolution inputs, 
by ensuring predictions are computed on semantically meaningful patches.
Crucially, we adopt this approach to avoid resizing the original images. 
Downscaling can introduce severe distortions and significantly degrade fine details, 
making small or distant objects unrecognizable. By processing the image in patches 
that match the ViT encoder's expected input size, we preserve spatial fidelity and 
maintain the integrity of small-scale structures, which is essential for 
accurate predictions.

In the zero-shot segmentation setting, we extend predicted probabilities to support multi-label 
segmentation through a dedicated post-processing pipeline and segmentation refinement, 
described in \cref{sec:post_processing}. This includes cosine blending of 
overlapping patch predictions, reconstruction of full-resolution probability 
maps via normalized weighted averaging, and a final refinement step using Conditional Random Fields (CRF).

The inference pipeline operates on the following inputs and component:
\begin{enumerate}
  \item Left image $I_L \in \mathbb{R}^{H \times W \times 3}$,
  \item Right image $I_R \in \mathbb{R}^{H \times W \times 3}$,
  \item SENSE-352 or SENSE-512 model, which computes predictions $f(i^{i}_L, i^{i}_R, j) \rightarrow P^{i}_j$,
  where $i \in {1, \dots, N}$ indexes patches from the sliding-window approach and $j \in {1, \dots, N_{\text{CLS}}}$ 
  denotes user-defined open-vocabulary text queries,
  \item Text queries corresponding to dataset classes~\cite{Cordts2016Cityscapes, Alhaija2018IJCV}.
\end{enumerate}
\vspace{1em}
The overall pipeline is illustrated in \cref{fig:pre_and_post_processing}.

\subsubsection{Stereo Images Pre-Processing}
\label{sec:pre_processing}
\begin{figure*}[t]
  \centering
  \includegraphics[width=\textwidth]{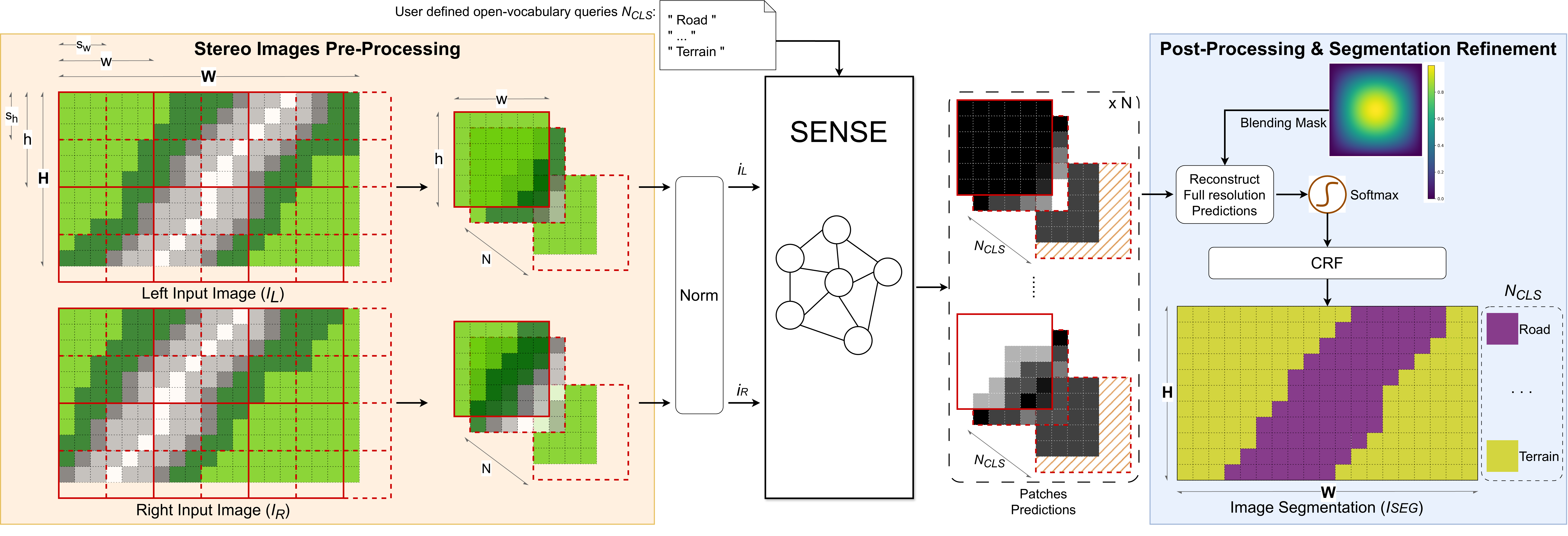}
  \caption{\textbf{Overview of the inference pipeline for large image resolutions.} \textit{Left}: overlapping 
  patch extraction from stereo input images ($I_L, I_R$) using a sliding-window strategy. 
  \textit{Middle}: patch-wise predictions for user-defined text queries ($N_{CLS}$) computed by SENSE 
  after normalization. 
  \textit{Right}: post-processing with cosine blending mask, reconstruction of full-resolution probability maps, 
  softmax normalization, and CRF-based segmentation refinement to produce the final segmentation map $I_{SEG}$.}
  \vspace{-1em}
  \label{fig:pre_and_post_processing}
\end{figure*}

Given stereo image pairs $(I_L, I_R) \in \mathbb{R}^{H \times W \times 3}$, as shown in \cref{fig:pre_and_post_processing} (left),
we extract overlapping patches of size $h \times w$ with strides $s_h = h/2$ and $s_w = w/2$. 
For each patch $i$:
\begin{equation}
  \begin{aligned}
  i^{i}_L &= I_L[y:y+h,; x:x+w], \\
  i^{i}_R &= I_R[y:y+h,; x:x+w].
  \end{aligned}
  \label{eq:patching}
\end{equation}

This yields a set of overlapping patches ${(i^{i}_L, i^{i}_R) \in \mathbb{R}^{h \times w \times 3}}\text{, with }{i=1,...,N}$, 
that capture semantic details across the stereo pair. 
Each patch undergoes image normalization before being processed by 
SENSE. This strategy ensures full coverage while maintaining computational efficiency by operating on 
smaller, semantically coherent regions.

\subsubsection{Post-Processing and Segmentation Refinement}
\label{sec:post_processing}
\cref{fig:pre_and_post_processing} (right) depicts the post-processing and segmentation refinement steps.
For each patch $i$ and text query $j$, the model outputs:
\begin{equation}
  P^{i}_j = f(i^{i}_L, i^{i}_R, j),
  \label{eq:patch_pred}
\end{equation}
where $P^{i}_j \in \mathbb{R}^{h \times w}$ is a 2D probability map ($352 \times 352$ for 
SENSE-352 or $512 \times 512$ for SENSE-512).

To merge overlapping predictions, we apply a cosine blending mask $M$ that emphasizes 
central regions and smoothly decays toward borders, reducing seam artifacts. 
For each text query $j$, we maintain:
\begin{enumerate}
  \item Prediction accumulator:
  \begin{equation}
    A_P(y:y+h, x:x+w) \mathrel{+}= P^i_j \odot M,
  \end{equation}
  \item Weight accumulator:
  \begin{equation}
    A_W(y:y+h, x:x+w) \mathrel{+}= M.
  \end{equation}
\end{enumerate}
The full-resolution probability map is then:
\begin{equation}
\widehat{P}_j = \frac{A_P}{A_W}.
\end{equation}
Finally, we stack ${\widehat{P}_j}\text{, with }{j=1,...,N_{CLS}}$, to 
form $\widehat{P} \in \mathbb{R}^{H \times W \times N_{\text{CLS}}}$, 
apply softmax across channels dimension, and refine using Conditional 
Random Field (CRF) technique, yielding the final segmentation $I_{\text{SEG}} \in \mathbb{R}^{H \times W \times N_{CLS}}$.

\subsection{Additional Qualitative Results}

\begin{figure*}
  \centering
  \includegraphics[width=\textwidth]{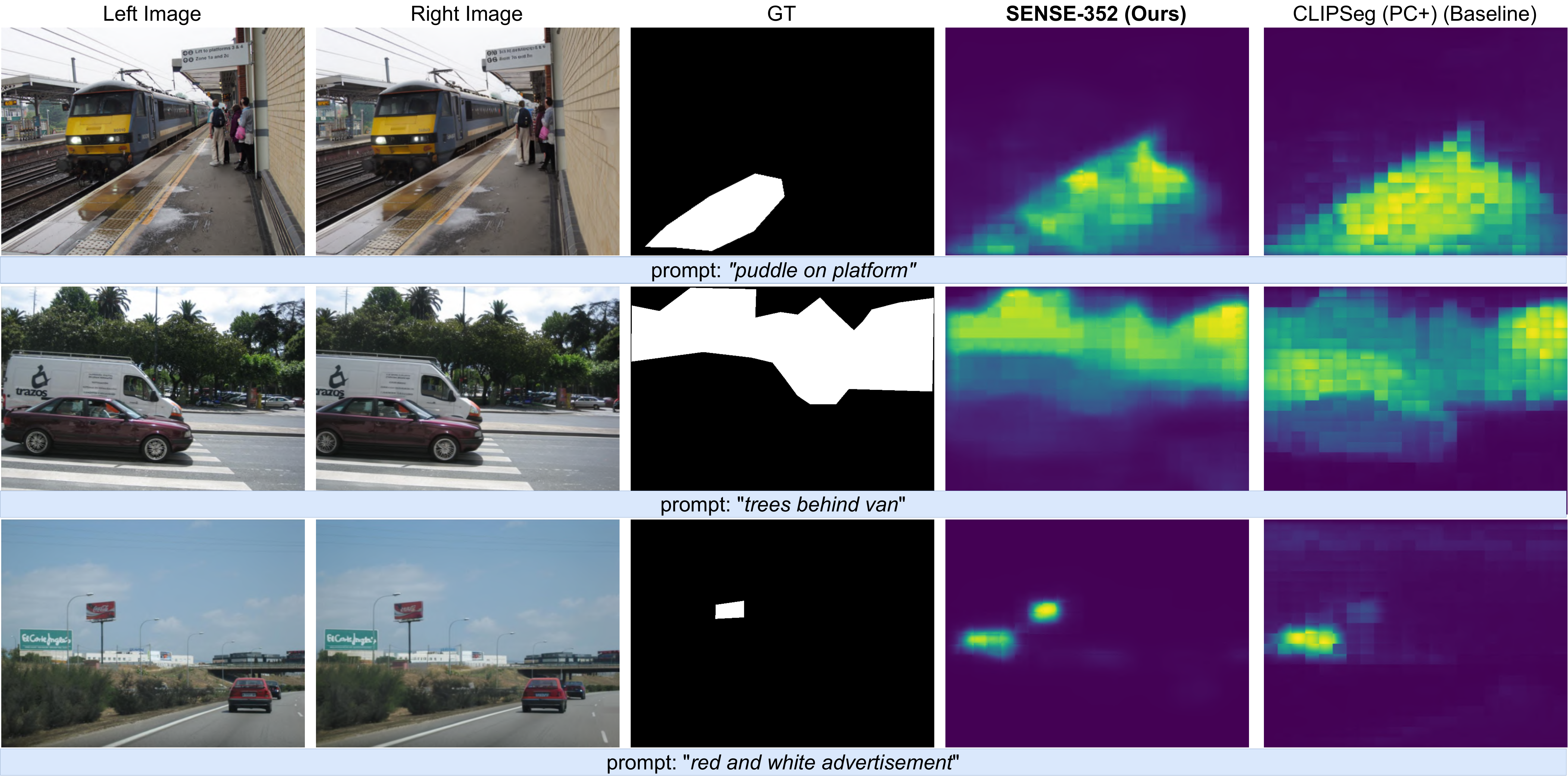}
  \caption{\textbf{Additional qualitative comparison of referring expression semantic segmentation} of SENSE-352 in PhraseStereo dataset
  with CLIPSeg (PC+) [24] method. We show results for diverse expressions, including object-specific (\textit{"puddle on platform"}), 
  relational (\textit{"trees behind van"}), and attribute-based (\textit{"red and white advertisement"}) prompts. 
  Predictions are visualized as sigmoid probability maps, highlighting confidence for the queried text prompt (blue box).}
  \vspace{-1em}
  \label{fig:qual_result_ref_exp_seg_supp}
\end{figure*}

\paragraph{Referring Expression Segmentation.}
This section presents additional qualitative results for the referring expression 
segmentation task, comparing our approach (SENSE-352) with the baseline CLIPSeg~\cite{luddecke2022image}. 
We include diverse examples covering simple and complex expressions, such as 
object-specific conditions (e.g., \textit{"puddle on platform"}), relational cues 
(e.g., \textit{"trees behind van"}), and attribute-based references (e.g., \textit{"red and white advertisement"}).
The visualizations show that SENSE-352 produces masks that are more accurate and 
spatially coherent with the ground truth, particularly in challenging scenarios 
involving multiple similar objects. In contrast, CLIPSeg often exhibits imprecise 
boundaries or wrong predictions that fail to fully capture the intended referent, 
even when the overall prediction appears plausible. 
\cref{fig:qual_result_ref_exp_seg_supp} illustrates representative comparisons, 
highlighting the improved grounding capability and segmentation precision achieved 
by our method.

\begin{figure*}[h]
  \centering
  \includegraphics[width=\textwidth]{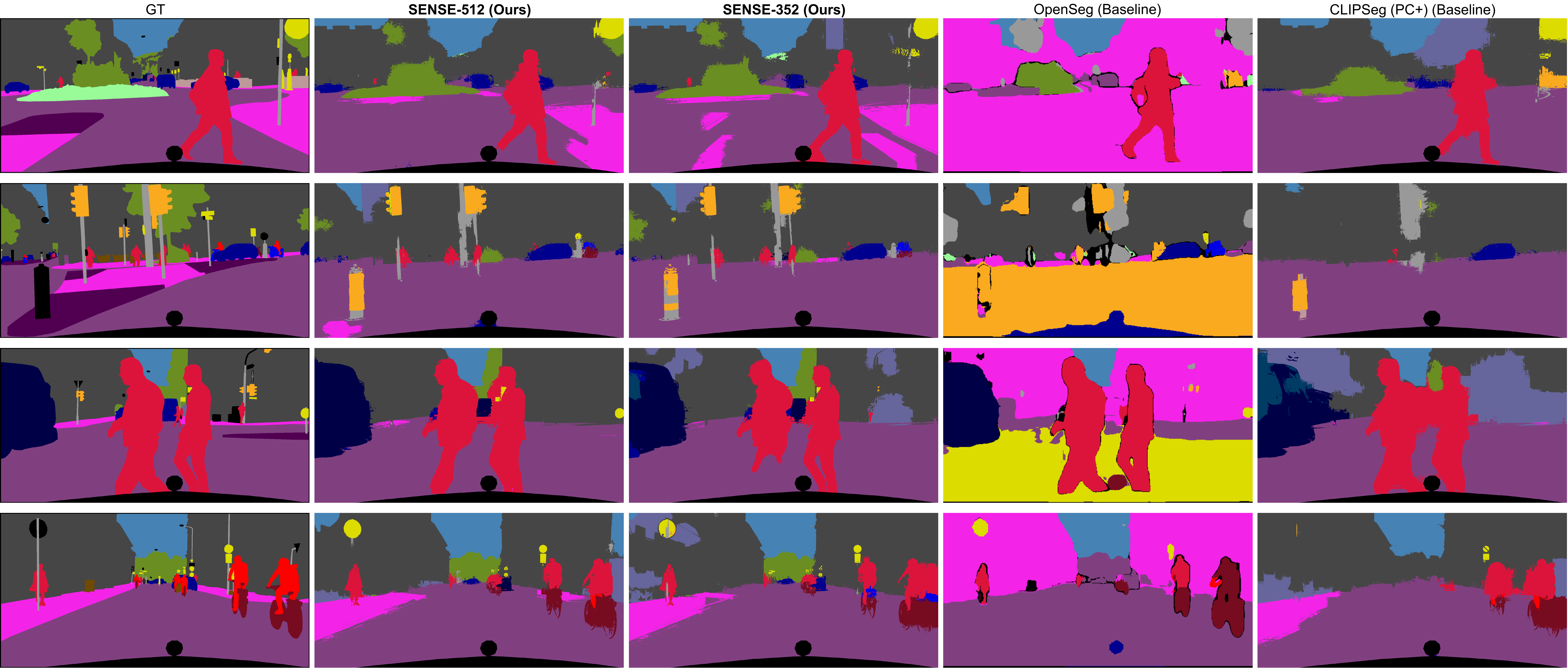}
  \caption{\textbf{Qualitative comparison of zero-shot semantic segmentation on Cityscapes.} We show predictions 
  from our models (SENSE-512 and SENSE-352) with baselines OpenSeg~\cite{ghiasi2022scaling} and CLIPSeg (PC+)~\cite{luddecke2022image}.}
  \vspace{-1.2em}
  \label{fig:qual_resut_zero_shot_supp}
\end{figure*}

\begin{figure*}[h]
  \centering
  \includegraphics[width=\textwidth]{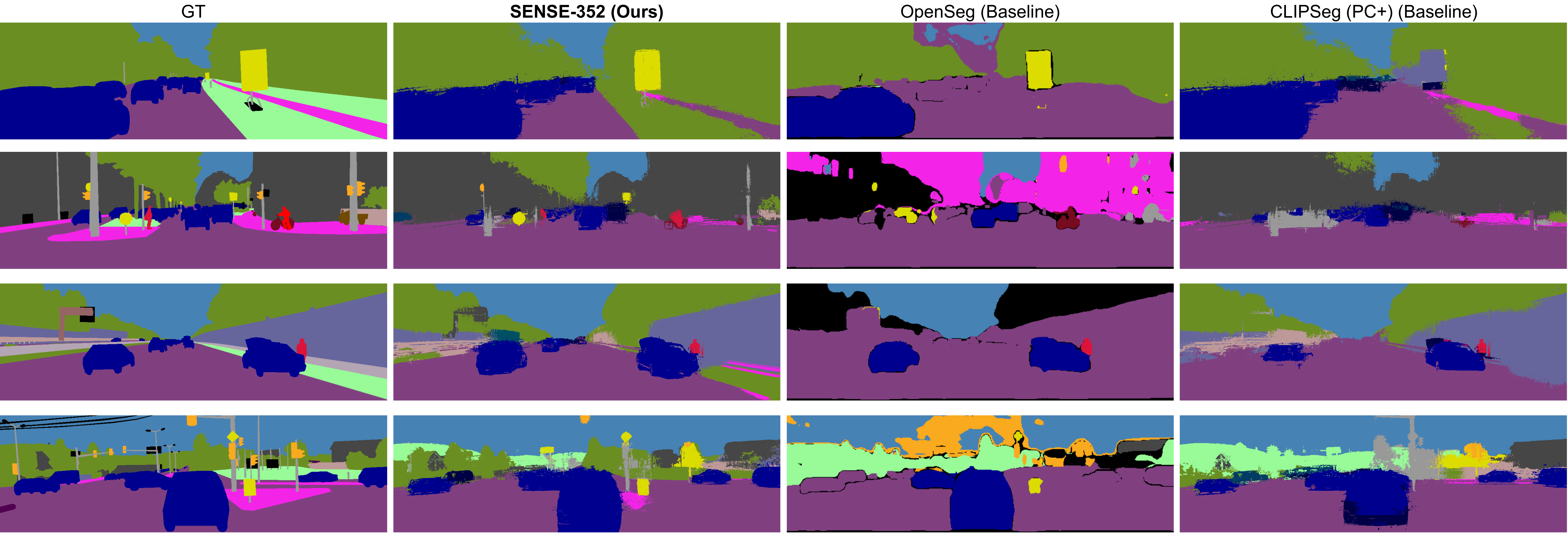}
  \caption{\textbf{Qualitative comparison of zero-shot semantic segmentation on KITTI 2015.} We show predictions 
  from our method (SENSE-352) and baselines OpenSeg~\cite{ghiasi2022scaling} and CLIPSeg (PC+)~\cite{luddecke2022image}.}
  \label{fig:qual_result_zero_shot_kitti_supp}
\end{figure*}

\vspace{-1em}
\paragraph{Zero-shot semantic segmentation.}
This section provides extended qualitative comparisons of zero-shot semantic 
segmentation on Cityscapes and KITTI. We visualize predictions from our method 
(SENSE-512 and SENSE-352) along with two strong zero-shot baselines, OpenSeg~\cite{ghiasi2022scaling} and 
CLIPSeg~\cite{luddecke2022image}. Representative scenes include diverse semantic concepts such as dynamic 
traffic participants (e.g., car, person, rider) and static structural 
elements (e.g., building, road, vegetation), under varying lighting and occlusion 
conditions.
As shown in \cref{fig:qual_resut_zero_shot_supp} and \cref{fig:qual_result_zero_shot_kitti_supp}, we provide qualitative comparisons 
on Cityscapes~\cite{Cordts2016Cityscapes} and KITTI 2015~\cite{Alhaija2018IJCV}, respectively. 
For both datasets, we follow the standard benchmark protocols. We use the official class names from~\cite{Cordts2016Cityscapes} 
as text prompts and evaluate on the test splits, ensuring consistent color mappings and fair comparison.
For Cityscapes, we compare SENSE-512, SENSE-352, OpenSeg~\cite{ghiasi2022scaling}, and CLIPSeg (PC+)~\cite{luddecke2022image}. 
The results show that SENSE-512 consistently produces sharper boundaries and more 
coherent regions than SENSE-352, demonstrating the benefit of larger stereo input 
sizes for capturing richer spatial context. OpenSeg captures high-level semantics, 
but struggles with fine-grained boundaries and small objects. CLIPSeg exhibits strong 
semantic segmentation recall, but it lacks spatial precision. For example, 
in the last row of \cref{fig:qual_resut_zero_shot_supp}, the person occluded on the left, partially hidden behind the traffic sign pole, is not predicted, illustrating its difficulty in handling details of occlusions.

In KITTI 2015, we compare SENSE-352 with OpenSeg~\cite{ghiasi2022scaling} and CLIPSeg (PC+)~\cite{luddecke2022image}. 
SENSE-352 delivers more accurate and spatially consistent predictions than both baselines, 
particularly for small or distant objects.
Notably, both SENSE variants, combined with our sliding-window strategy and CRF-based 
pre- and post-processing (\cref{sec:pre_post_processing}), significantly improve segmentation quality compared to 
CLIPSeg. For completeness, baseline methods do not use our sliding-window strategy or post-processing, as they rely on their own pre- and post-processing steps. 
CLIPSeg~\cite{luddecke2022image} is evaluated with its refinement module, while OpenSeg~\cite{ghiasi2022scaling} and the other baselines operate as multiclass models. 
In contrast, SENSE performs binary, text-conditioned inference, requiring a different processing pipeline.
These qualitative results complement the quantitative metrics reported in the main paper 
and further emphasize the improved results achieved by our approach.

\begin{table*}[t]
\centering
\begin{tabular}{l  @{\,\qquad\,} l  @{\,\qquad\,} c @{\,\qquad\,} c c c}
\toprule
Method & Disparity Estimation Model & Time (ms) & mIoU & IoU$_{\text{FG}}$ & AP \\
\midrule
\multirow{3}{*}{\textbf{SENSE-352} (Ours)}
& Selective-IGEV$^\dagger$~\cite{wang2024selective}                 & 194.74$^\dagger$ & \textbf{47.2}$^\dagger$ & 57.0$^\dagger$ & 78.8$^\dagger$ \\
& HITNet~\cite{tankovich2021hitnet}                         & 104.09 & 47.1 & 56.8 & 78.3 \\
& MobileStereoNet~\cite{shamsafar2022mobilestereonet} &  \textbf{76.91} & 47.1 & \textbf{57.1} & \textbf{78.9} \\
\midrule
\multirow{3}{*}{\textbf{SENSE-512} (Ours)}
& Selective-IGEV$^\dagger$~\cite{wang2024selective}                 & 284.54$^\dagger$ & \textbf{46.5}$^\dagger$ & 56.3$^\dagger$ & 77.7$^\dagger$ \\
& HITNet~\cite{tankovich2021hitnet}                         & 136.01 & 46.4 & \textbf{56.4} & \textbf{77.8} \\
& MobileStereoNet~\cite{shamsafar2022mobilestereonet} &  \textbf{91.68} & 46.3 & 56.3 & \textbf{77.8} \\
\bottomrule
\end{tabular}
\caption{\textbf{SENSE's computational time with different stereo matching models.}
We report inference time in milliseconds and performance on the Referring Expression Segmentation task.
Best results in \textbf{bold}. $^\dagger$ indicates the model and results discussed in \cref{tab:ref_expression_seg} and \cref{tab:runtime_comparison}.}
\label{tab:computation_requirements_supp_models}
\vspace{0.5em}
\end{table*}

\subsection{Descriptions as text queries}

As shown in \cref{tab:cityscapes_classes_decriptions}, the Cityscapes~\cite{Cordts2016Cityscapes} dataset 
provides detailed descriptions for each semantic class. These descriptions 
demonstrate how natural language can serve as a concise and expressive representation 
of scene semantics. 

\begin{figure*}[h]
  \centering
  \includegraphics[width=\textwidth]{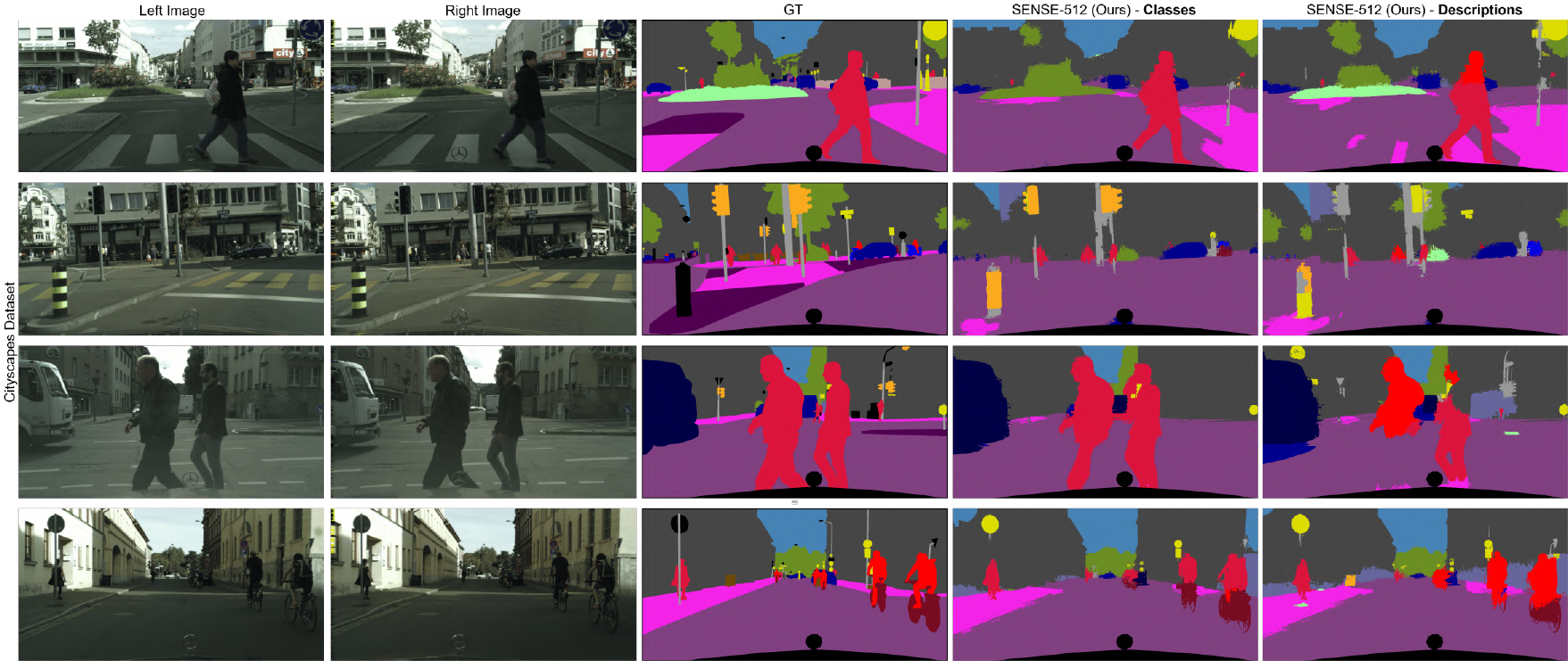}
  \caption{\textbf{From class labels to class descriptions on Cityscapes.} We compare SENSE-512 predictions when prompted with standard \textbf{class} names and the corresponding natural language \textbf{descriptions} shown in \cref{tab:cityscapes_classes_decriptions}. Despite using descriptive text instead of fixed labels, SENSE-512 achieves segmentation quality comparable to ground truth, demonstrating its ability to interpret scenes through concise language prompts while maintaining spatial consistency across stereo views.}
  \vspace{-1em}
  \label{fig:qual_res_descriptions}
\end{figure*}

Rather than relying on class labels, in this section we evaluate SENSE using these textual descriptions 
as prompts for open-vocabulary segmentation. This enables SENSE to interpret and 
segment scenes through language-based queries that capture richer context than simple labels. 
For example, instead of the single word \textit{"car"}, the corresponding description conveys 
both object identity and contextual attributes. In \cref{fig:qual_res_descriptions}, we see a 
small effect on performance for descriptive prompt forms, while SENSE-512 achieves segmentation performance still 
comparable to its class-based results, highlighting its ability to generalize to natural language queries.

By combining stereo cues with natural language conditioning, 
SENSE provides a flexible pathway for scene understanding. We believe that this 
capability supports complex reasoning and decision-making, allowing perception systems 
to connect semantic interpretation with planning.

\subsection{Computation requirements}
As observed in \cref{sec:computation_time} of the main paper, we highlighted that the disparity estimation module dominates runtime and that, without it, 
SENSE becomes substantially faster than all compared baselines. In particular, in the maain paper, \cref{tab:runtime_comparison} shows that OpenSeg~\cite{ghiasi2022scaling} was the 
fastest baseline, running at 123.35 ms. In this section, we further analyze our methods, SENSE, with alternative stereo matching models. 
We report inference time using an NVIDIA GeForce RTX 3090 Ti GPU, for a stereo image pair and one text query.
As shown in \cref{tab:computation_requirements_supp_models}, replacing Selective-IGEV~\cite{wang2024selective} with lighter stereo matchers significantly reduces inference time. For SENSE-352, HITNet~\cite{tankovich2021hitnet} reduces runtime to 104.09 ms, while 
MobileStereoNet~\cite{shamsafar2022mobilestereonet} achieves the fastest performance at 76.91 ms, substantially outperforming OpenSeg. A similar trend appears at $512 \times 512$ resolution, where MobileStereoNet again provides the 
largest speedup (91.68 ms compared to 284.54 ms for Selective-IGEV). Importantly, despite these large runtime improvements, the Referring Expression 
Segmentation metrics remain closely aligned with those obtained using Selective-IGEV, with only small variations, confirming that segmentation quality is 
robust to the choice of stereo matcher.
These findings reinforce that the disparity estimator in SENSE is fully plug-and-play (as described in \cref{sec:SDAF_module}). Although the model is trained using 
Selective-IGEV, different stereo modules can be used at inference without retraining. In particular, replacing Selective-IGEV with a real-time stereo 
depth estimation method such as MobileStereoNet~\cite{shamsafar2022mobilestereonet} makes SENSE substantially more suitable for edge and embedded deployment, 
while preserving segmentation accuracy.

\paragraph{Model Size.} 
Both SENSE-512 and SENSE-352 process the stereo pair using the CLIP ViT-B/16 image encoder with shared weights, 
applied separately to the left and right images ($\approx$ 86.2M parameters in total), together with a CLIP text encoder 
($\approx$ 37.8M parameters).
When stereo disparity is included, the decoder together with the SIEF and SDAF modules adds 
approximately 3.12 M parameters. In the configuration without stereo matching (SDAF removed), 
the decoder with SIEF contains $\approx$ 2.01 M parameters for both variants of SENSE.

Regarding FLOPs, we report a direct comparison with OpenSeg, which requires $\approx$ 170 GFLOPs. 
In contrast, SENSE-512 has approximately 190 GFLOPs, arising from applying the shared-weight CLIP ViT-B/16 encoder 
twice ($\approx$ 92 GFLOPs per forward pass), together with the CLIP text encoder ($\approx$ 4 GFLOPs) and our decoder ($\approx$ 3 GFLOPs).
These results demonstrate that SENSE maintains competitive computational efficiency, providing faster inference than OpenSeg~\cite{ghiasi2022scaling} while supporting open-vocabulary, 
stereo-aware segmentation.

\begin{table*}[t]
\centering
\small
\begin{tabular}{lp{15cm}}
\toprule
\textbf{Class} & \textbf{Description} \\
\midrule
road & Part of ground on which cars usually drive, i.e. directions, streets. Including the markings on the road. Areas only delimited by markings from the main road (no texture change) are also road, e.g. bicycle lanes, roundabout lanes, or parking spaces. \\
sidewalk & Part of ground designated for pedestrians or cyclists. Delimited from the road by some obstacle, e.g. curbs or poles (might be small), not only by markings. Often located at the sides of a road. \\
building & Building, skyscraper, house, bus stop building, garage, car port. If a building has a glass wall that you can see through, the wall is still building. \\
wall & Individual standing wall. Not part of a building. \\
fence & Fence including any holes. \\
pole & Small mainly vertically oriented pole. E.g. sign pole, traffic light poles. If the pole has a horizontal part this part is also considered pole. If there are things mounted at the pole that are neither traffic light nor traffic sign, then these things might also be labeled pole. \\
traffic light & The traffic light box without its poles. \\
traffic sign & Sign installed from the state/city authority, usually for information in an everyday traffic scene, e.g. traffic signs, parking signs, direction signs without their poles. No ads/commercial signs. \\
vegetation & Tree, hedge, all kinds of vertical vegetation. Plants attached to buildings are usually not annotated separately and labeled building as well. \\
terrain & Grass, all kinds of horizontal vegetation, soil or sand. These areas are not meant to be driven on. \\
sky & Open sky, without leaves of tree. Includes thin electrical wires in front of the sky. \\
person & A human that would walk, it's person. People walking, standing or sitting on the ground, on a bench, on a chair. This also includes toddlers, someone pushing a bicycle or standing next to it with both legs on the same side of the bicycle. \\
rider & Riders or drivers as person of bicycle, motorbike, scooter, skateboards, horses, roller-blades, wheel-chairs, road cleaning cars, cars without roof. \\
car & Car, jeep, SUV, van with continuous body shape, caravan, no other trailers. \\
truck & Truck, box truck, pickup truck. Including their trailers. \\
bus & Bus for persons, public transport or long distance transport. \\
train & Vehicle on rails, tram, train. \\
motorcycle & Motorbike, moped, scooter without the driver. \\
bicycle & Bicycle without the driver. \\
\bottomrule
\end{tabular}
\caption{Cityscapes semantic classes and their descriptions.}
\label{tab:cityscapes_classes_decriptions}
\end{table*}

\paragraph{Acknowledgements}
This work was supported by the CIFRE program (Convention Industrielle de Formation par la Recherche) under ANRT (Association Nationale de la Recherche et de la Technologie) grant No. 2024/0356, in collaboration with Centre Inria d’Universite C$\hat{o}$te d’Azur, France, and NXP Semiconductors.